\documentclass[preprint,12pt]{elsarticle}

\usepackage{CJKutf8}
\usepackage{float}
\usepackage{caption}
\usepackage{enumitem}
\setlist[itemize]{leftmargin=*}
\usepackage{amsmath}
\usepackage{url}
\usepackage{tikz}
\usetikzlibrary{positioning}
\usepackage{algorithm}
\usepackage{algpseudocode}

\usepackage{array}
\usepackage{booktabs} %调整表格线与上下内容的间隔
\usepackage{multirow}

\usepackage{graphicx}
\usepackage{tikz}
\usepackage{forest}
\usetikzlibrary{trees,positioning,shapes,shadows,arrows.meta}

\usepackage{amsthm}
\newtheorem{remark}{Remark}

\usepackage{amssymb}
\journal{Journal of Biomedical Informatics}

\begin{document}

\begin{frontmatter}



\title{Data Augmentation Techniques for Chinese Disease Name Normalization}


\author[1]{\mbox{Wenqian Cui}}
\author[2]{\mbox{Xiangling Fu}}
\author[2]{\mbox{Shaohui Liu}}
\author[2]{\mbox{Mingjun Gu}}
\author[3]{\mbox{Xien Liu}}
\author[3]{\mbox{Ji Wu}}
\author[1]{\mbox{Irwin King}}

\affiliation[1]{organization={The Chinese University of Hong Kong}, country={Hong Kong}}
\affiliation[2]{organization={Beijing University of Posts and Telecommunications}, country={Beijing}}
\affiliation[3]{organization={Tsinghua University}, country={Beijing}}

\begin{abstract}
Disease name normalization is an important task in the medical domain. It classifies disease names written in various formats into standardized names, serving as a fundamental component in smart healthcare systems for various disease-related functions. Nevertheless, the most significant obstacle to existing disease name normalization systems is the severe shortage of training data. Consequently, we present a novel data augmentation approach that includes a series of data augmentation techniques and some supporting modules to help mitigate the problem. 
Our proposed methods rely on the Structural Invariance property of disease names and the Hierarchy property of the disease classification system. The goal is to equip the models with extensive understanding of the disease names and the hierarchical structure of the disease name classification system.
Through extensive experimentation, we illustrate that our proposed approach exhibits significant performance improvements across various baseline models and training objectives, particularly in scenarios with limited training data.
\end{abstract}

\begin{graphicalabstract}
\begin{figure}[ht]
\centering
\includegraphics[width=1.0\textwidth]{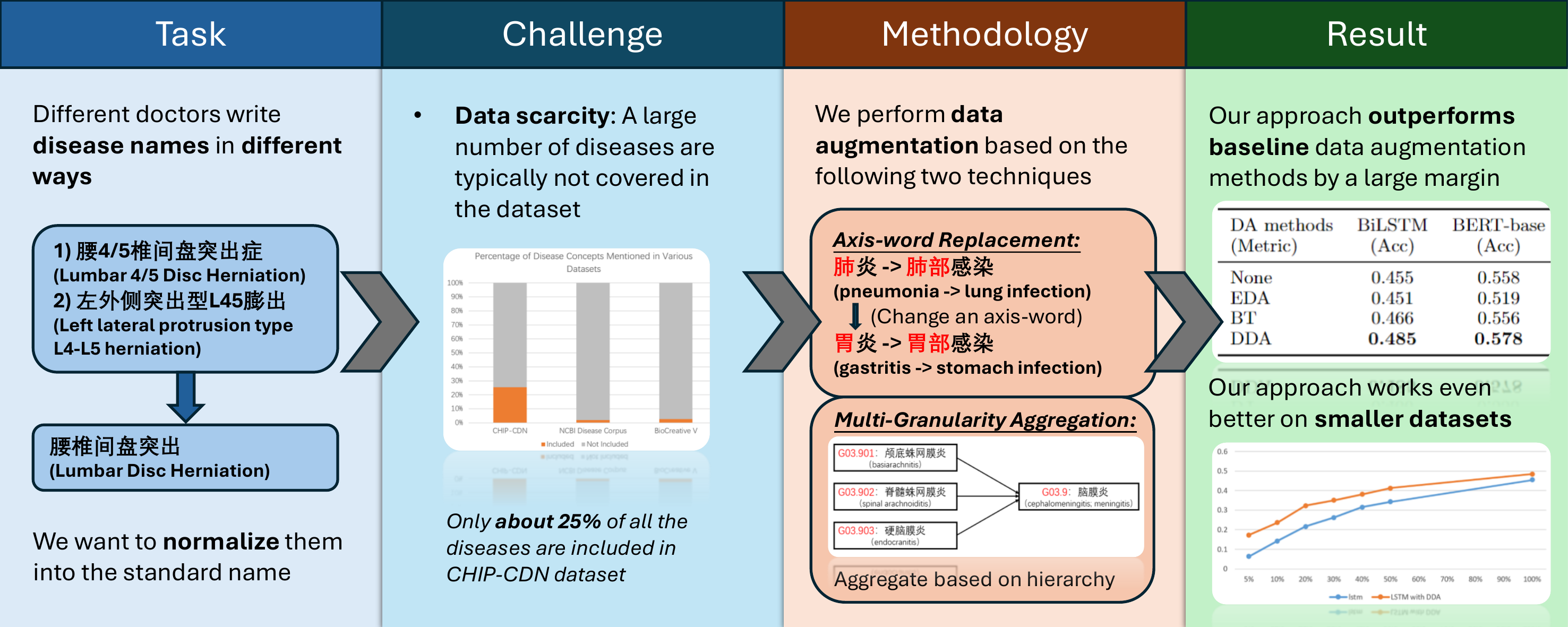}
% \caption{Caption of the image}
% \label{fig:graphical_abstract}
\end{figure}
\end{graphicalabstract}

\footnotetext{
The non-standard abbreviations used in this paper:

\begin{itemize}
    \item DDA: Disease Data Augmentation
    \item AR: Axis-word Replacement
    \item MGA: Multi-Granularity Aggregation
    \item ngm: $n$-gram matching score
    \item UDN: Unnormalized Disease Name
    \item SDN: Standard Disease Name
    \item NoDC: Number of Disease Concepts
    \item EDA: Easy Data Augmentation
    \item BT: Back Translation
\end{itemize}
}

\begin{highlights}
\item Disease name normalization is a crucial task that classifies disease names written in diverse formats into standard names.
\item The primary challenge in disease name normalization lies in the scarcity of annotated training data, hindering the development of robust models.
\item Our novel data augmentation approach, Disease Data Augmentation (DDA), focuses on manipulating both the key elements (axis words) and the hierarchical structures of the disease names.
\item Regular data augmentation methods fail to perform well on the disease name normalization task.
\item Our proposed DDA approach demonstrates remarkable efficacy in enhancing the performance of disease normalization tasks across various baseline models.
\item Particularly effective in scenarios with smaller datasets, our DDA approach achieves impressive results, recovering nearly 80\% of the full performance for specific evaluation metrics.
\item By striking a perfect balance between model complexity and performance, our DDA approach outperforms various Large Language Model (LLM) baselines, showcasing its efficiency and effectiveness.
\end{highlights}


\end{frontmatter}


\section{Introduction}
\begin{figure}[t]
  \centering
  \includegraphics[width=1.0\textwidth]{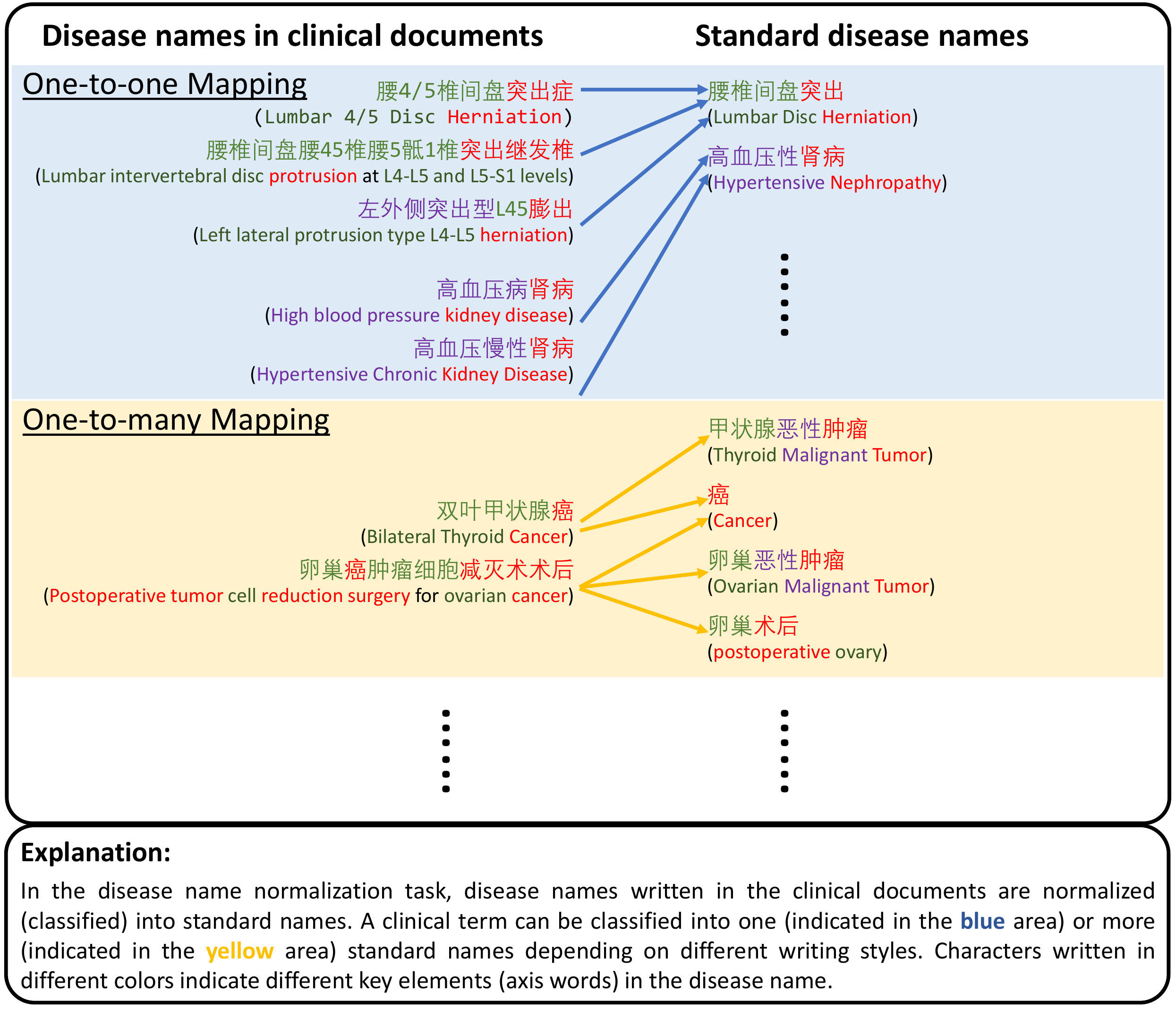}
  \caption{Examples and illustrations of the disease name normalization task}
  \label{fig:task_process}
\end{figure}

Disease names play a pivotal role in modern intelligent healthcare systems as it is involved in diverse tasks such as intelligent consultation \cite{chipcdn_kdd}, auxiliary diagnosis \cite{auxiliary_diagnosis1, auxiliary_diagnosis2}, automated International Classification of Diseases (ICD) coding \cite{automatic_icd_coding1, automatic_icd_coding2, automatic_icd_coding3}, Diagnosis-Related Groups prediction \cite{drg1, drg2, drg3}, etc. However, in clinical settings, doctors often write disease names according to their own habits and preferences, leading to numerous variations for the same disease. Therefore, to carry out additional operations on disease names, it is necessary to normalize them into standard names. As a result, disease name normalization, which entails classifying the diagnosis terms in clinical documents to standard names or classifications, plays a critical role in the ecosystem. Figure \ref{fig:task_process} illustrates the disease name normalization task.

% There are three main challenges in the Chinese disease name normalization task.
% First, \textbf{varied writing styles}. Disease names can vary widely due to the different writing habits of doctors. Some doctors adhere strictly to the standard disease name, while others prefer to use abbreviated versions or even create personalized disease names based on the patient's symptoms.
% Second, \textbf{semantic density}. Disease names are typically short but carry significant semantic information. Even with numerous shared characters, diseases can still have vastly different meanings, and a single character change can lead to a dramatic shift in semantic meaning. For instance, ``\begin{CJK*}{UTF8}{gbsn}髂总动脉夹层\end{CJK*} (Common iliac artery dissection)" is a disease from the upper half of the human body while ``\begin{CJK*}{UTF8}{gbsn}颈总动脉夹层\end{CJK*} (Common carotid artery dissection)" is from the lower half, although they share almost all the characters.
% Third, \textbf{data scarcity}: A large number of diseases are typically not covered in the training set, leading to few-shot or zero-shot scenarios. In CHIP-CDN dataset, only about 25\% of all the diseases are provided, and the scarcity is even more pronounced in NCBI Disease Corpus and BioCreative V datasets as indicated in Figure \ref{data_scarcity}. In this case, it is nearly impossible for the models to gain comprehensive knowledge about the disease system.
% % 可以把pubmed什么的也加上，一起画一张大图

\begin{figure}[t]% [htbp]
    \centering
    \includegraphics[width=1.0\textwidth]{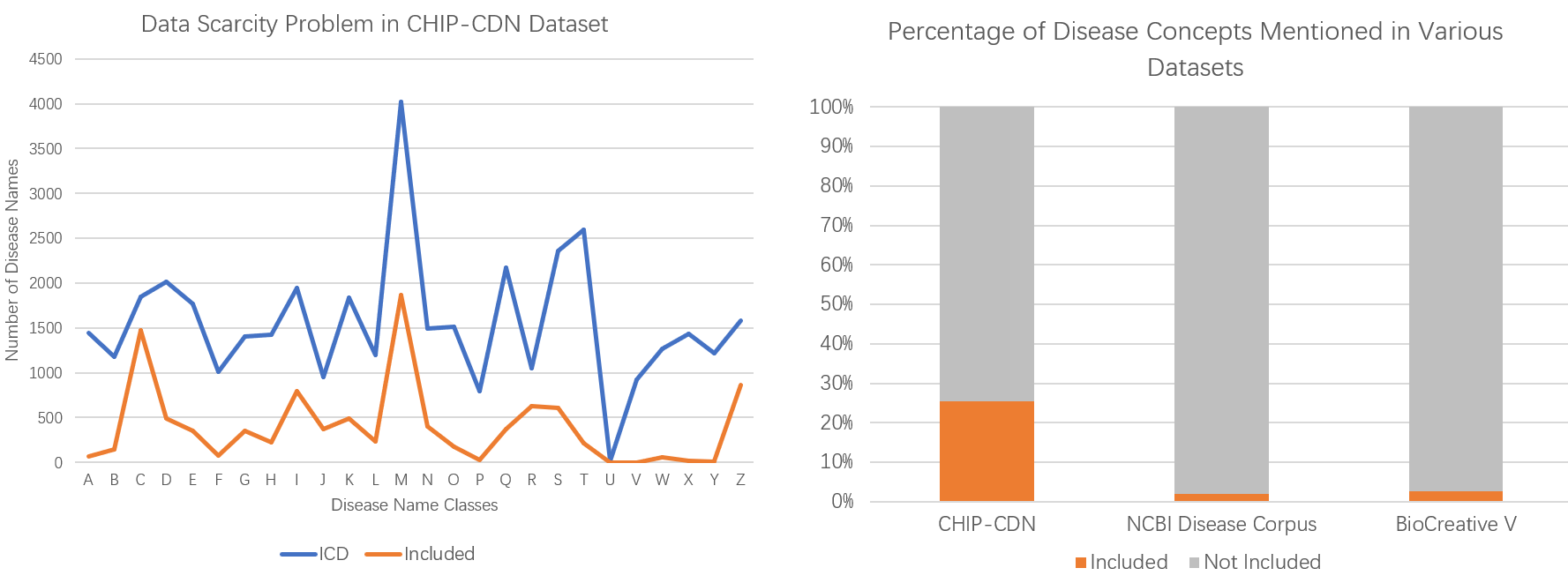}
    \caption{Data scarcity problem in commonly-used disease name normalization and related datasets. Left: The number of disease names presented in the CHIP-CDN training set versus the total number of disease names classified by the first letter. Right: The percentage of disease concepts mentioned in various datasets.
    % The blue line represents the overall amount of diseases in ICD coding classified by the first coding letter, and the red line represents the number of diseases provided by the CHIP-CDN training set.
    }
    \label{data_scarcity}
\end{figure}

One of the main challenges in the disease normalization task is data scarcity. Specifically, a substantial proportion of disease names and concepts are typically not covered in the training set, leading to few-shot or zero-shot scenarios in the normalization process. For example, in CHIP-CDN dataset \cite{zhang2021cblue}, only about 25\% of all the diseases are provided, and the scarcity is even more pronounced in NCBI Disease Corpus \cite{ncbi_disease_corpus} and BioCreative V \cite{li2016biocreative_dataset} datasets as indicated in Figure \ref{data_scarcity}. In this case, it is extremely difficult for the models to gain comprehensive knowledge about the disease system. Although collecting more data seems to be a natural solution to address this challenge, it is more difficult to perform in the medical field due to privacy concerns and the requirement for expertise. Hence, in this work, we utilize data augmentation as a workaround to address the data scarcity problem.

% figure1：按照icd首字母，分别标出一共有的编码数，和CDN训练集中的数量
% 写出zero-shot和少于等于2个的个数

% % 提一下DA分类，以及我们的方法属于哪一类
% Among the challenges mentioned above, data scarcity is the most significant one, as machine learning models heavily depend on the availability of sufficient training data. However, collecting and labeling data in the medical field poses greater difficulty due to privacy concerns and the requirement for expertise. In this work, we utilize data augmentation as a workaround to address the data scarcity problem.

We design a data augmentation approach including a set of data augmentation methods and some supporting modules for Chinese disease name normalization tasks called Disease Data Augmentation (DDA). Our data augmentation methods are based on the following two characteristics of the disease names. 
Firstly, disease names exhibit \textbf{Structural Invariance}. Disease names consist of various key elements (axis words) such as anatomical region, clinical manifestations, etiology, and pathology. When replacing one of the elements with another of the same category, it will typically still result in a valid disease name. For example, when the anatomical region ``\begin{CJK*}{UTF8}{gbsn}髂\end{CJK*} (Iliac)" of the disease ``\begin{CJK*}{UTF8}{gbsn}髂总动脉夹层\end{CJK*} (Common iliac artery dissection)" is replaced by another region ``\begin{CJK*}{UTF8}{gbsn}颈\end{CJK*} (Carotid)", we derive a name with the same type of disease but locates in another region ``\begin{CJK*}{UTF8}{gbsn}颈总动脉夹层"\end{CJK*} (Common carotid artery dissection)". Therefore, in disease name normalization, we can create new training data by replacing specific elements in pairs of clinical and standard disease names simultaneously. 
Secondly, the classification system of disease names demonstrates \textbf{Hierarchy} property, allowing for more specified descriptions to be encompassed into larger, more coarse groups. For instance, the more detailed disease definition, ``\begin{CJK*}{UTF8}{gbsn}急性喉炎\end{CJK*} (Acute Laryngitis)", can also be viewed as ``\begin{CJK*}{UTF8}{gbsn}喉炎\end{CJK*} (Laryngitis)". Hence, we can augment the training data by assigning the label of a fine-grained disease to its father disease in the classification system. 
By augmenting the training data under the above rules, we provide the models with an extensive understanding of disease names, particularly those that are absent in the original training set. Additionally, our methods enhance the models' comprehension of the hierarchical structure and the relationships among different disease names.

Our experiments demonstrate that our DDA approach outperforms all other data augmentation counterparts and effectively enhances the performance of various disease name normalization baselines. Furthermore, our approach can perform much better with smaller datasets and can achieve nearly 80\% of the full performance even when no data from the training set is provided.

The \textbf{Statement of Significance} of this paper is as follows.
\begin{itemize}
    \item \textbf{Problem or Issue:} The primary challenge in disease name normalization, which involves classifying variously formatted disease names into standardized terms, is the scarcity of annotated training data. This hinders the development of effective and robust models.
    \item \textbf{What is Already Known:} Disease names have two main characteristics: they are composed of several key elements, and their classification system exhibits a hierarchical structure.
    \item \textbf{What this Paper Adds:} This paper introduces a novel data augmentation approach named Disease Data Augmentation (DDA), which leverages the two previously mentioned characteristics of disease names. We demonstrate through experiments that DDA significantly enhances the performance of the Chinese disease name normalization task compared to baseline approaches and across various backbone models.
\end{itemize}

% To summarize our contributions.
% \begin{itemize}
%     \item We introduce a set of data augmentation techniques for the Chinese disease normalization task based on the two characteristics of disease names.
%     \item We design a novel data augmentation approach encompassing the data augmentation techniques as well as some supporting modules.
%     \item We conduct extensive experiments to show that our data augmentation approach effectively improves the performance of the Chinese disease name normalization task compared to baseline approaches and across various backbone models.
% \end{itemize}

The subsequent sections of this paper are structured as follows. Section 2 provides an in-depth exploration of the background of the disease name normalization task and discusses related work pertinent to our research. In Section 3, we present a detailed explanation of our methodology. Section 4 outlines the experiments conducted, details the dataset utilized, and presents the results obtained. The final sections conclude the paper and discuss the limitations and future works of the study.

% This is because general data augmentation methods have the potential to alter the meaning of disease names significantly. For example, if we apply random deletion \cite{wei2019eda} to ``\begin{CJK*}{UTF8}{gbsn}阻塞性睡眠呼吸暂停\end{CJK*} (Obstructive Sleep Apnoea)", it will result in ``\begin{CJK*}{UTF8}{gbsn}阻塞性睡眠\end{CJK*} (Obstructive Sleep)", which represents a completely different disease. However, we design our data augmentation methods to tailor to disease names, which are classified as sampling-based methods in \cite{DA_survey}, making it suitable for creating stronger applications.

\label{}

\section{Background and Related Work}

\begin{figure}
    \centering
    
    % I am using only the basic, xnode, and tnode styles
    \tikzset{
        basic/.style  = {draw, text width=3cm, align=center, font=\sffamily\small, rectangle},
        root/.style   = {basic, rounded corners=2pt, thin, align=center, fill=green!30},
        onode/.style = {basic, thin, rounded corners=2pt, align=center, fill=green!60,text width=3cm,},
        tnode/.style = {basic, thin, align=left, fill=pink!60, text width=10em, align=center, font=\sffamily\scriptsize}, % , font=\sffamily\tiny
        xnode/.style = {basic, thin, rounded corners=2pt, align=center, fill=blue!20,text width=4cm,},
        wnode/.style = {basic, thin, align=left, fill=pink!10!blue!80!red!10, text width=6.5em},
        edge from parent/.style={draw=black, edge from parent fork right}
    }
    
    \begin{forest} for tree={
        grow=east,
        growth parent anchor=west,
        parent anchor=east,
        child anchor=west,
        edge path={\noexpand\path[\forestoption{edge},->, >={latex}] 
             (!u.parent anchor) -- +(10pt,0pt) |-  (.child anchor) 
             \forestoption{edge label};}
        % font=\small % Set the font size for all nodes in the tree
    }
    % l sep is used for arrow distance
    [Biomedical Entity Linking, basic,  l sep=7mm,
        [Other Biomedical Entity Linking Tasks, xnode, l sep=5mm,
            % [Shaikh et al.\, 2008, tnode]
            % [Bhuyan et al.\, 2011, tnode]
            % [Bou et al.\, 2013, tnode] 
            ]
        [Disease Name Normalization, xnode, l sep=5mm,
            [Language Model-based Approach: \cite{zhu2023promptcblue,liu2023moelora}, tnode]
            [\textbf{Data Augmentation: Ours}, tnode] 
            [Task-specific Model Architecture: \cite{chipcdn_kdd,zhang2021cblue,disambiguation_GNN,chipcdn_infusing_bert}, tnode] 
            ]
        [Disease Entity Linking, xnode, l sep=5mm,
            [Language Model-based Approach: \cite{DEL_DN_LLM}, tnode]
            [Machine Learning-based Approach: \cite{leaman2013dnorm,lou2020investigating_DN,DN_with_graph_embeddings,medical_DN,bern2_DN,DN_ACL_2020,deep_learning_DN,wright2019normco}, tnode]
            [Rule-based Approach: \cite{dogan2012inference_method_DN}, tnode] 
            ] 
        ]
    \end{forest}

    \caption{A taxonomy of biomedical entity linking methods. Our approach falls into the data augmentation category within the disease name normalization task.}
    \label{fig:taxonomy}
\end{figure}
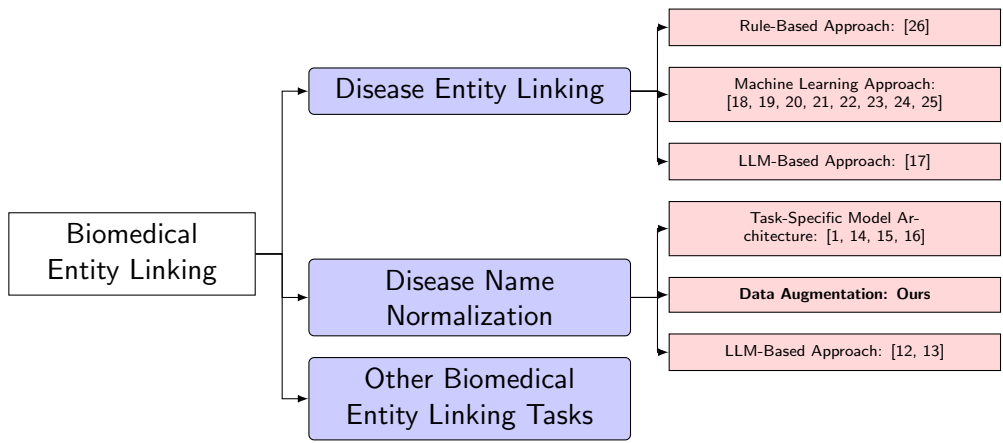

\subsection{Disease Name Normalization}
Disease name normalization refers to the process of matching or classifying disease terms written by doctors in clinical documents to their standard names based on certain classification systems. The term ``Disease name normalization" is widely used in literature, but its meaning varies and can refer to different tasks. While some literature defines disease name normalization as the process of retrieving and matching disease terms in lengthy medical description texts, such as medical literature abstracts in the NCBI Disease Corpus \cite{ncbi_disease_corpus} and BioCreative-V-CDR-Corpus \cite{li2016biocreative_dataset} datasets, we argue that this task does not align well with the name ``normalization" because the unnormalized and normalized entity should fall into a same concept, such as disease name. The task that retrieves from lengthy medical description texts should be categorized as Disease Entity Linking. This falls under the broader category of Biomedical Entity Linking (BEL), which, according to \cite{BEL_overview}, is described as ``the task of mapping of spans of text within biomedical documents to normalized, unique identifiers within an ontology". We see Disease Entity Linking as an end-to-end approach to classify disease names mentioned in the description text, and this larger task (Disease Entity Linking) can be divided into two subtasks: identifying disease-related corpora in the lengthy description text and normalizing the identified corpora into standard names based on the classification system. In this work, we define the disease name normalization task as the second subtask of the Disease Entity Linking task, which is consistent with the definitions in \cite{chipcdn_kdd,zhu2023promptcblue,zhang2021cblue}, and we use the 10\textsuperscript{th} version of the ICD system as the standard classification system. Figure \ref{fig:taxonomy} shows the taxonomy of the Biomedical Entity Linking task, including our work and the related works.

\subsection{Data Augmentation on Text Data} % 找一些22，23年的DA methods
Data augmentation is a technique that generates new data from existing datasets to increase data volume and help prevent model overfitting. While it is simpler to augment image data without losing meaning, text augmentation is more challenging due to its unstructured nature \cite{ng2020ssmba}. Some approaches, like those suggested by \cite{wei2019eda}, apply character-level modifications such as replacement, insertion, swap, and deletion, though these can introduce grammatical errors. Back translation \cite{understanding_back_translation} maintains semantic integrity but lacks diversity and depends on the quality of the translation tools.

There are also more complex methods used for text augmentation. \cite{kim2022alp} uses lexicalized probabilistic context-free grammars to capture the complex structure of natural language and replace words, resulting in effective results. However, this grammar-based approach is challenging to apply to specialized domains like medicine. Pre-trained language models are also used for augmentation; for example, \cite{ng2020ssmba} and \cite{wu2019conditional} utilize the MLM objective in BERT \cite{wu2019conditional} to regenerate masked words, and \cite{kumar2020pretraincomparison} compare different pre-trained model methods. However, these methods can alter the original text's meaning after several MLM replacements.
Additionally, Semi-supervised learning can also augment data using the vast amount of unlabeled data. \cite{berthelot2019mixmatch} use MixUp to guess low-entropy labels of augmented data and combine labeled and unlabeled data to derive a loss term, while \cite{xie2020uda} perform data augmentation on unlabeled data for consistency training. However, our focus here is solely on the labeled data rather than the unlabeled data. For an extensive overview of text data augmentation methods, we refer the readers to \cite{DA_survey}.

\subsection{Data augmentation on medical data}
While the majority of studies focus on the impact of data augmentation on general text data, some studies explore the potential of data augmentation operations on medical text data. Several works concentrate on synonym replacement in medical terms. \cite{falis2022icd9dataaug} and \cite{abdollahi2021zapaidataaug} utilize the Unified Medical Language System (UMLS) \cite{umls_system} to identify medical synonyms for replacements in classification texts. \cite{falis2022icd9dataaug} also replaces both medical terms in raw texts and the classification label to generate new training data, focusing on the ICD-coding task. While their work mainly centers on replacing the entire medical term, we investigate the possibility of replacing the components within the medical terms. Furthermore, \cite{ansari2021mentalhealth} examines the performance of EDA, conditional pre-trained language models, and back translation for data augmentation on social media texts for mental health classification. \cite{wang2020stsdataaug} proposes Segment Reordering as a data augmentation technique to preserve the semantic meaning of medical texts. \cite{wang2020stspesudo} use pre-trained language models fine-tuned on General Semantic Textual Similarity (STS-G) data to generate pseudo-labels on medical STS data and then undergo iterative training.

\section{Proposed Methodology}
\begin{figure}[htb]
    \centering
    \includegraphics[width=1.0\textwidth]{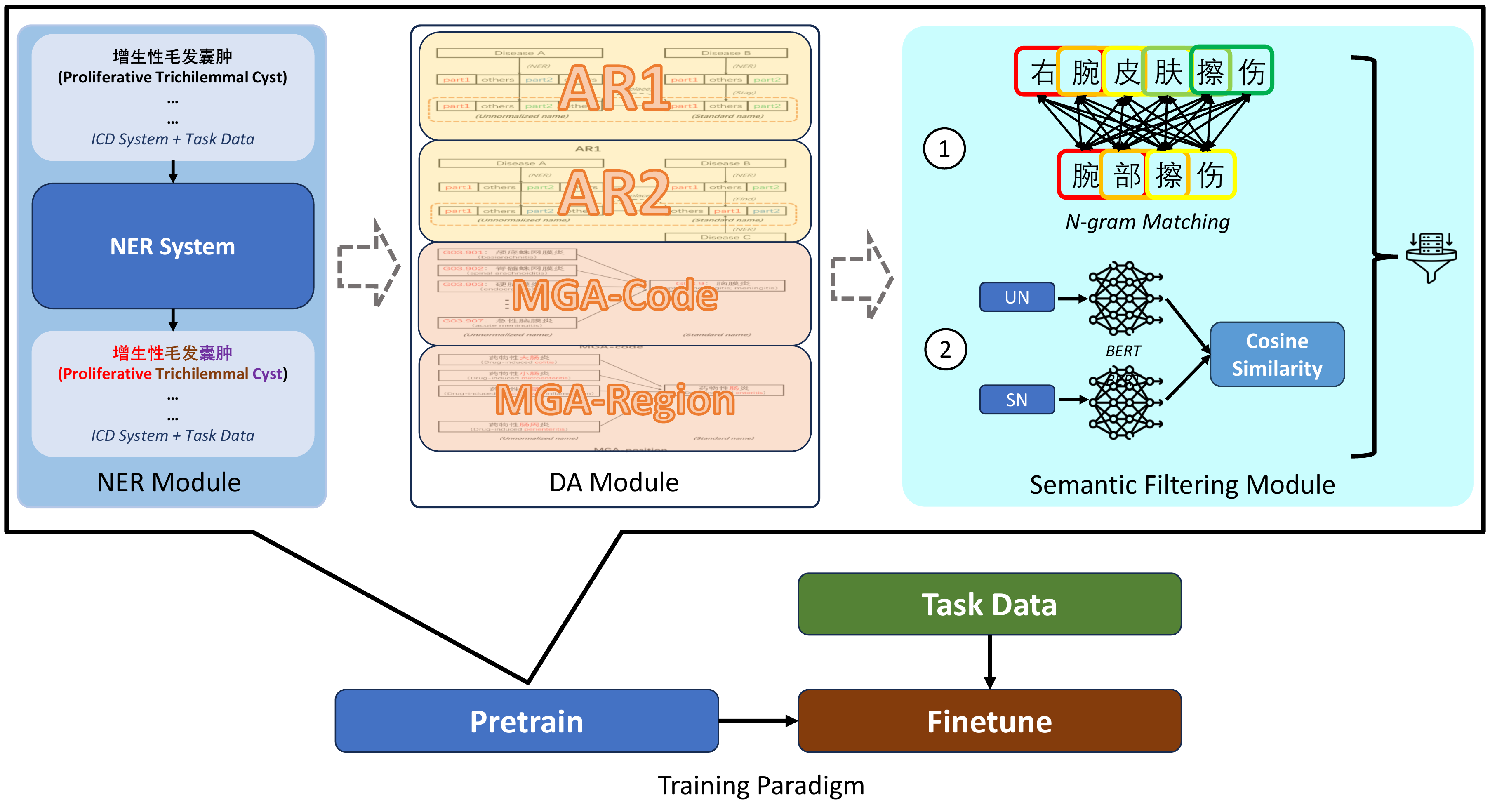}
    \caption{The overall pipeline of our proposed Disease Name Normalization (DDA) approach. AR1, AR2, MGA-Code, and MGA-Region are the four proposed data augmentation techniques, and their details are illustrated in Figure \ref{methods}.}
    \label{overall_pipeline}
\end{figure}

This section introduces the details of the pipeline for our proposed data augmentation approach called Disease Data Augmentation (DDA), as depicted in Figure \ref{overall_pipeline}. Our approach consists of three main components: 1) A named entity recognition (NER) module, 2) a data augmentation module, and 3) a semantic filtering module. Specifically, all the inputs will first go through a NER system to locate and identify all the elements, and then the results are sent to the data augmentation (DA) module to generate new pairs of data. A semantic filtering module is at the end to filter out unwanted pairs.

We first define the concept of ``axis word" and the type of axis words used in this work. We then introduce the three main modules of our approach. Finally, we illustrate the training paradigm of our approach. For clarification, we use the terms ``unnormalized disease name" and ``standard disease name" to denote the input and output of the disease normalization system, respectively.

\subsection{Axis Word}
The disease names are composed of several elements (axis words), which include but are not limited to etiology, pathology, clinical manifestations, anatomical region, chronicity, degree type, characteristic, etc. \cite{axisword1, axisword2}. 
Therefore, we define axis words as the elements within the disease names. For ease of expression, we merge etiology and pathology into disease center and select from all remaining elements into three main categories: disease center, anatomical region, and disease characteristic. With these three axis words, a large portion of disease names can be combined. Table \ref{tab:axis-words} shows the definition of the three axis words with an example alongside them.

\begin{table}[t]
\fontsize{10}{11}\selectfont
\captionsetup{skip=6pt}
\centering
\caption{Definition and an example of the axis words used in this work.}
\begin{tabular}{@{}c p{4cm} p{5.5cm}@{}}
\toprule
Axis Word & Definition & Example \\
\midrule
Disease Center & The minimal term that describes the nature of a disease, which may include etiology and pathology. It defines the main category of the disease. & \begin{CJK*}{UTF8}{gbsn}增生性毛发\textbf{\underline{囊肿}}\end{CJK*}\newline (Proliferative Trichilemmal \textbf{\underline{Cyst}})\\
Anatomical Region & A part of the human body that has actual meaning in anatomy. This part of the disease name indicates which part of the human body is ill. & \begin{CJK*}{UTF8}{gbsn}增生性\textbf{\underline{毛发}}囊肿\end{CJK*} \newline (Proliferative \textbf{\underline{Trichilemmal}} Cyst)\\
Disease Characteristic & The characteristic of a disease that indicate the subtype or the cause of the disease. & \begin{CJK*}{UTF8}{gbsn}\textbf{\underline{增生性}}毛发囊肿\end{CJK*} \newline (\textbf{\underline{Proliferative}} Trichilemmal Cyst)\\
% \midrule
% Reason & x & x \\
% Symptom & x & x \\
% Population & x & x \\
% Medical Tests & x & x \\
% Degree & x & x \\
\bottomrule
\end{tabular}
\label{tab:axis-words}
\end{table}

\subsection{Named Entity Recognition Module}
The first module of our approach is a named entity recognition (NER) system to locate and identify the axis words of all the input disease names.
To build the NER system, we select 5,000 diseases from ICD system \cite{icd10} based on its taxonomy and ask doctors to annotate the labels (i.e., the three axis words) in BIO format \cite{bio_tagging_ner}. 
% \footnote{BIO format indices entities in sequences by labeling ``beginning", ``inside", and ``outside".}. 
We use the traditional ``BiLSTM + CRF" as the NER model architecture. Specifically, there are three BiLSTM layers \cite{lstm} with a hidden dimension of 100, a fully connected layer, and a CRF layer \cite{crf}. The model achieves 0.794 of micro F1 score in our final evaluation. 
% \begin{remark}
% Although the NER tool is crucial to the whole system, it is not the focus of this work. It is just for the use of locating axis words, and it can be replaced by any module that can achieve the same function.
% \end{remark}

\subsection{Data Augmentation Module}
\begin{figure}[t]
    \centering
    \includegraphics[width=1.0\textwidth]{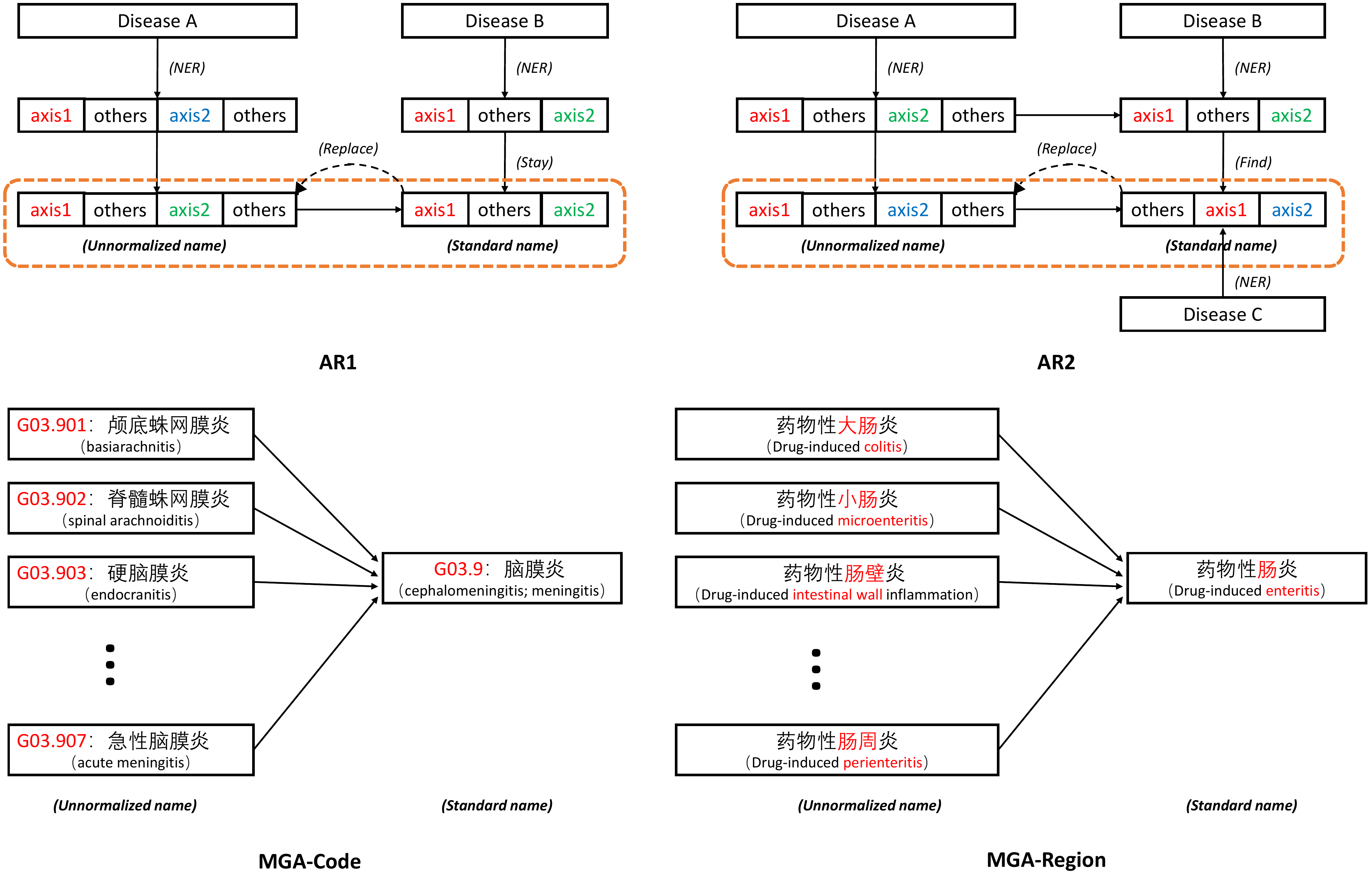}
    \caption{Illustration of our proposed data augmentation techniques. The upper portion of the figure depicts the Axis-word Replacement methods, and the lower portion depicts the Multi-Granularity Aggregation methods.}
    \label{methods}
\end{figure}

The data augmentation modules consist of four data augmentation methods, and they are divided into two main categories: Axis-word Replacement (AR) and Multi-Granularity Aggregation (MGA). The main purpose of our data augmentation methods is to provide the model with additional knowledge, so we focus on exploring the components and relationships within diseases to give the model a comprehensive understanding of the various components and the hierarchical classification system of disease names. Figure \ref{methods} illustrates the two categories and four types of data augmentation methods\footnote{We will open source the augmentation code and the augmented result on Github.}. We also present pseudo-code for all the four data augmentation methods in \ref{apdx:pseudo-code}.

\subsubsection{Axis-word Replacement (AR)}
Axis-word Replacement method is designed based on the assumption that disease names exhibit \textbf{Structural Invariance} property. This means that replacing an axis word in a disease name with another word of the same type still results in a meaningful disease name. Since there are often matches of axis words between an unnormalized disease name and a standard disease name in the disease name normalization task, simultaneously replacing the same axis word in both the unnormalized name and the standard name can typically ensure that the newly generated pair will still match. We leverage both the ICD and task data (data from the disease name normalization training set) to perform Axis-word Replacement. The detailed descriptions of each category of Axis-word Replacements are as follows: 
\begin{itemize}
\item \textbf{AR1}: AR1 is illustrated in the top left corner of Figure \ref{methods}. First, we select a pair of diseases (disease A and disease B) that share one or more axis words (axis1 in the figure) but differ in another axis word (axis2 in the figure). Then, we replace axis2 in disease A with the same axis2 in disease B. 
(Note: disease A can be chosen from any sources, but disease B can only be chosen from the standard ICD system as it serves as the label of a disease name normalization pair.)
Algorithm \ref{algorithm:AR1} provides the pseudo-code for AR1.

We can choose either of the three axis words to perform replacement:
\begin{itemize} % 这三个AR1的子类，可以文字描述？整体的AR1用伪代码表示
\item AR1 - Region: Perform AR1 by fixing the disease center and replacing the anatomical region.
\item AR1 - Center: Perform AR1 by fixing the anatomical region and replacing the disease center.
\item AR1 - Characteristic: Perform AR1 by fixing both the disease center and the anatomical region and replacing the disease characteristic.
\end{itemize}
\item \textbf{AR2}: AR2 is illustrated in the top right corner of Figure \ref{methods}. First, we select a pair of unnormalized-standard diseases from the disease name normalization training set. Let the unnormalized disease be disease A, and the standard disease be disease B. Then, find disease C from the ICD system that shares one or more axis words (axis1 in the figure) but differ in another axis word (axis2). Finally, we replace axis2 in disease A to be the same axis2 in disease C, so that the replaced disease A and disease C can form a new disease name normalization pair. Algorithm \ref{algorithm:AR2} provides the pseudo-code for AR2.

Similarly, we can choose either of the three axis words to perform replacement:
\begin{itemize}
\item AR2 - Region: Perform AR2 by fixing the disease center and replacing the anatomical region.
\item AR2 - Center: Perform AR2 by fixing the anatomical region and replacing the disease center.
\item AR2 - Characteristic: Perform AR2 by fixing both the disease center and the anatomical region and replacing the disease characteristic.
\end{itemize}

\end{itemize}

% 上下部位，解释一下为什么上下位部位要加在里面，再说一下部位树，还有伪代码，讲一下生成的逻辑
% 提一下输入增强和输出增强的概念
% 关系化表示：A->B->C等

\subsubsection{Multi-Granularity Aggregation (MGA)}
Multi-Granularity Aggregation (MGA) method is designed based on the \textbf{hierarchical} structure of the ICD system, and the granularity levels of the structure is organized by the length of the ICD codes. For example, in ICD-10 Beijing Clinical Version 601, the disease name of the four-digit code ``A18.2" is ``\begin{CJK*}{UTF8}{gbsn}外周结核性淋巴结炎\end{CJK*} (Peripheral Tuberculous Lymphadenitis)", and it has in total 10 child diseases that have a fine-grained description, with six-digit codes ranging from ``A18.201" to ``A18.210", such as ``A18.201: \begin{CJK*}{UTF8}{gbsn}腹股沟淋巴结结核\end{CJK*} (Inguinal lymph node tuberculosis)" and ``A18.202: \begin{CJK*}{UTF8}{gbsn}颌下淋巴结结核\end{CJK*} (Submandibular lymph node tuberculosis)". This shows that the ICD system exhibits a tree-like structure, where a coarse-defined disease can be associated with multiple fine-grained diseases.

The granularity levels of the hierarchical structure include the first 3, 4, and 6-digit codes. For example, \begin{CJK*}{UTF8}{gbsn}``K81.0: 急性胆囊炎(Acute Cholecystitis)"\end{CJK*} and \begin{CJK*}{UTF8}{gbsn}``K81.1: 慢性胆囊炎(Chronic Cholecystitis)"\end{CJK*} share the first 3-digit code but differ in the 4th-digit code. As a result, they are both from the Cholecystitis category but differ in the type of the disease.
We observe that the meaning between disease names that share the first 3-digit code but differ in the 4th-digit code can be quite distinct, but the meaning would be much more similar if the disease names share the first 4-digit code. Therefore, We implement MGA augmentation using the following methods:

\begin{itemize}
\item \textbf{MGA - Code}: We assign the label of a 6-digit disease name to its corresponding 4-digit disease name. We refer to this method as ``aggregation" because typically a 4-digit disease name can be linked to several 6-digit disease names, allowing the model to learn which diseases are similar. MGA-code is depicted in the bottom left part of Figure \ref{methods}. Algorithm \ref{algorithm:MGA-Code} provides the pseudo-code for MGA - Code.

We can perform aggregation using disease names from different sources:
\begin{itemize}
\item MGA - Code 1: The 6-digit diseases are obtained from the ICD system.
\item MGA - Code 2: The 6-digit diseases are obtained from the diseases in the task training set with 6-digit ICD disease labels.
\end{itemize}

% 这里还是提一下position tree的事情
\item \textbf{MGA - Region}: In addition to the ICD system, anatomical regions also exhibit a tree-like hierarchical structure, where smaller regions can be grouped together to form a larger region. We identify disease names that share the same center but where the region of one disease is the larger region of another. We then assign the classification labels of the smaller-region disease names with their corresponding larger-region disease names. The MGA-Region method is depicted in the bottom right part of Figure \ref{methods}. In this method, the larger-region disease names, serving as the standard names, must be sourced from the standard ICD system. Algorithm \ref{algorithm:MGA-Position} provides the pseudo-code for MGA - Region.

Similarly, we can perform aggregation using disease names from different sources:
\begin{itemize}
\item MGA - Region 1: The lower region disease names are obtained from the ICD system.
\item MGA - Region 2: The lower region disease names are obtained from the names in the task training set.
\end{itemize}
\end{itemize}

\begin{remark}
In the human body, a region is considered the larger-region in relation to another if it covers a larger area. To determine the larger or smaller regions of a region, we create an expert-annotated region tree document that organizes anatomical regions into a tree data structure. This region tree is used to identify upper and lower relations. Similar results can be obtained using other sources containing knowledge bases of human anatomy.
\end{remark}

\begin{remark}
Before performing the data augmentation methods, we exclude all the diseases having an ICD code starting with P, Q, and any letter after T, for those diseases are mainly related to pregnancy, giving birth, and long description texts, which we found do not follow the above assumptions we made. We then perform the four data augmentation methods on the remaining disease names to form the augmented dataset.
\end{remark}

\subsection{Semantic Filtering Module}
We design a filtering module to remove generated disease pairs with low confidence. As discussed in previous sections, we perform data augmentation by replacing the axis words within a disease name (AR) and manipulating matching relationships by aggregation (MGA). However, although our methods follow the nature and characteristics of the disease names, replacing an axis word of an unnormalized disease with another does not always result in an authentic disease, and the aggregation operation is not always accurate. Therefore, to ensure the quality of the generated data, we propose a semantic filtering module as a post-processing step. Our filtering method is under the assumption that the unnormalized names should not deviate too much from the standard name, and if so, the generated disease names or the relationship have a high probability of being inauthentic.

% \begin{equation}
% \label{eq:n-gram}
% \text{ngm}(UDN, SDN) = \frac{{\sum_{i=1}^{\min(j, k)} \left| n\text{-gram}_i(UDN) \cap n\text{-gram}_i(SDN) \right|}}{{\min(j, k)}}, 
% \end{equation}

We measure the level of deviation or similarity in the semantic filtering module based on two criteria. The first part is a normalized $n$-gram matching (ngm) score between an unnormalized disease name (UDN) and a standard disease name (SDN): 
\begin{equation}
\label{eq:n-gram}
\text{ngm}(UDN, SDN) = \frac{{\sum_{n=1}^{\min(j, k)} \left| n\text{-gram}(UDN) \cap n\text{-gram}(SDN) \right|}}{{\min(j, k)}}, 
\end{equation}
where $j$ and $k$ are the lengths of UDN and SDN, respectively. 
Specifically, for each pair, we generate $n$-grams from $n$ equals 1 to the length of the shorter name in the pair. We then calculate the number of matched pairs and divide it by the length of the shorter name. This equation is utilized to penalize the pairs that have a large difference in length and do not share a fair amount of common characters. It measures the similarity in the character level.
The second part is a cosine similarity score between the contextual embeddings of UDN and SDN outputted by BERT \cite{devlin2018bert}, i.e., 
\begin{equation}
\label{eq:cosine_similarity}
% \text{{similarity}}(UDN, SDN) = \frac{{\text{{BERT}}(UDN) \cdot \text{{BERT}}(SDN)}}{{\|\text{{BERT}}(UDN)\| \|\text{{BERT}}(SDN)\|}}
\text{{similarity}}(UDN, SDN) = \text{{cosine}}(\text{{BERT}}(UDN), \text{{BERT}}(SDN)).
\end{equation}
It measures the similarity from the contextual semantic level.
The final dataset is derived by filtering out generated data pairs below the threshold of the normalized $n$-gram score or the cosine similarity score, 
\begin{equation}
\label{eq:overall_semantic_filtering}
\begin{aligned}
\text{{Final Dataset}} = \{(GeneratedPairs) | & \text{{ngm}}(UDN, SDN) > \alpha \\
& \land \text{{similarity}}(UDN, SDN) > \beta\}.
\end{aligned}
\end{equation}
where we set $\alpha$ and $\beta$ to be the threshold for the normalized $n$-gram score and the cosine similarity score, respectively. In this work, we set $\alpha$ and $\beta$ to be 0.7 and 0.8, respectively. The final number of paired disease names generated by each data augmentation method is shown in \ref{sec:DA result statistic}.

\subsection{Training Paradigm}
We train the models in a two-step fashion: the augmented data is used in the pre-training phase, and the original task data is then used in the fine-tuning phase. The reason is that although the semantic filtering module assists in eliminating fictitious disease names produced by the data augmentation module, it does not ensure that the remaining generated names are all genuine, and these fictitious disease names have the potential to negatively impact the overall task performance. 
Considering that our primary objective is to utilize a large volume of data to equip the model with extensive knowledge, we can leverage the generated disease pairs in the pre-training stage. After that, we finetune the models with the original task data to get the final results. Since we leverage several baselines to evaluate our approach, and they have different training objectives, we make the pre-training objective exactly the same as the fine-tuning objective for each baseline method.

\section{Experiments}
In this section, we conduct experiments to answer the following four research questions (RQs). 
\begin{itemize}
    \vspace{-2mm}
    \item RQ1: How does the proposed approach compare in effectiveness to different data augmentation baselines?
    \vspace{-3mm}
    \item RQ2: What is the individual contribution of each component in the proposed approach to the final outcome?
    \vspace{-3mm}
    \item RQ3: Considering our focus on tackling data scarcity, will the proposed approach demonstrate greater effectiveness on smaller datasets?
    \vspace{-3mm}
    \item RQ4: In the age of Large Language Models (LLMs), how does our proposed approach perform in comparison to LLM baselines?
\end{itemize}
% How does the proposed method compare in effectiveness to different baselines? What is the individual contribution of each component in the proposed method to the final outcome? Considering our focus on tackling data scarcity, will the proposed method demonstrate greater effectiveness on smaller datasets? Lastly, in the age of Large Language Models (LLMs), how does our proposed method perform in comparison to LLM baselines?

\subsection{Dataset}
\subsubsection{CHIP-CDN}
We evaluate the effectiveness of our data augmentation approach on a Chinese disease name normalization dataset called CHIP-CDN. CHIP-CDN originates in the CHIP-2019 competition and was collected in CBLUE\footnote{CBLUE: A Chinese Biomedical Language Understanding Evaluation Benchmark.} \cite{zhang2021cblue}. The dataset contains 6,000 unnormalized-standard disease pairs in the training set, 1,000 pairs in the validation set, and 2,000 pairs in the test set. In this dataset, the data pairs are not strictly a one-to-one mapping. Some unnormalized names are matched to several different standard names.

\subsubsection{Other Related Datasets}
We also try to find some English datasets to perform the experiments. In previous sections, we mentioned the inconsistency of the disease name normalization concept in various literature. Most of them use two main datasets to perform the task, namely NCBI Disease Corpus \cite{ncbi_disease_corpus} and BioCreative-V-CDR-Corpus \cite{li2016biocreative_dataset}. Both of them contain a certain number of PubMed abstracts written in English, and the task is to identify the disease concepts within the texts. However, as mentioned earlier, this task falls under the Biomedical Entity Linking rather than the Disease Name Normalization task, so we are not able to utilize them. Table \ref{tab:dataset_summary} summarizes the abovementioned datasets.

\begin{table}[H]
% \fontsize{9}{10}\selectfont
\captionsetup{skip=6pt}
\centering
\caption{Summary of different disease name normalization datasets, where ``Train/Val/Test" is the dataset split and ``NoDC" represents ``Number of Disease Concepts".}
\resizebox{\textwidth}{!}{%
\begin{tabular}{lcccc}
\toprule
Dataset Name & Language & Train/Val/Test & NoDC & Source \\
\midrule
CHIP-CDN & Chinese & 6,000/2,000/10,000 & 10,325 & Electronic Medical Records\\
NCBI Disease Corpus & English & 593/100/100 & 790 & PubMed Abstracts\\
BioCreative-V-CDR-Corpus & English & 500/500/500 & 1,082 & PubMed Articles\\
\bottomrule
\end{tabular}
}
\label{tab:dataset_summary}
\end{table}

\subsection{Experimental Setup}
We assess our approach using four baseline models: BiLSTM \cite{lstm}, BERT-base \cite{devlin2018bert}, CDN-Baseline (from CBLUE) \cite{zhang2021cblue}, and B\textsc{i-hard}NCE \cite{chipcdn_kdd}. For the BiLSTM model, we employ two BiLSTM layers with a hidden dimension of 256, followed by an MLP layer for classification. For the BERT-base model, we utilize the CLS vector \cite{devlin2018bert} within the BERT architecture for classification. CDN-Baseline is the baseline method presented in the CBLUE paper \cite{zhang2021cblue}, which introduces the CHIP-CDN dataset. It is based on BERT-base model and follows a ``recall-match" approach, where all relevant standard disease names for an unnormalized disease are recalled first, and the unnormalized disease is then matched to the final decision. B\textsc{i-hard}NCE is a contrastive learning-based method that has demonstrated effectiveness in symptom detection tasks and is also based on BERT-base model. It treats disease name normalization as a retrieval problem. The selection of these baseline models aims to showcase the effectiveness of our approach across different model types and training objectives. Specifically, we validate the effectiveness of DDA on both non-pretrained (BiLSTM) and pre-trained models (the other three), on models with simple (BERT-base) and complex (CDN-Baseline and B\textsc{i-hard}NCE) pre-training objectives.

We report all the metrics on the validation set. For the BiLSTM model and BERT-base model, we assess the model performance using accuracy. For these two models, we treat disease name normalization as a multi-class classification rather than a multi-label classification task. Therefore, if an unnormalized disease is matched to several standard diseases, we consider the data sample correctly predicted as long as one of the standard diseases is correctly predicted. We design the experiments in this way to simplify the model as much as possible and to more clearly illustrate the effectiveness of DDA. For CDN-Baseline, we adhere to the settings in CBLUE \cite{zhang2021cblue}, which uses F1 as the evaluation metric tailored to the multi-label setting. The F1 is calculated using precision (P) and recall (R), which are defined using the number of ``unnormalized-standard" disease pairs\footnote{If an unnormalized disease name is matched to three standard disease names, we say there are three disease pairs here.}:
\begin{equation}
\label{eq:f1}
P = \frac{k}{n}, R = \frac{k}{m}, F1 = \frac{2 \times P \times R}{P + R}, 
\end{equation}
where $m$ is the total number of data pairs in the evaluation dataset, $n$ is the number of predicted pairs, and $k$ is the number of correctly predicted pairs. As for B\textsc{i-hard}NCE, it is structured as a retrieval problem, so we report the RECALL@5 and NDCG@5 metrics following the original paper \cite{chipcdn_kdd}.

\subsection{Comparison with Different Data Augmentation Approaches (RQ1)}
We evaluate the effectiveness of our data augmentation approach by comparing it to two baseline approaches: EDA \cite{wei2019eda} and Back Translation (BT)\footnote{we use the youdao translation tool at https://fanyi.youdao.com/} \cite{understanding_back_translation}. We select EDA and BT as our benchmarks because they are commonly employed in various studies and represent the two primary categories of NLP data augmentation approaches---noise-based and paraphrase-based approaches---as outlined in \cite{DA_survey}. Furthermore, the decision is supported by the work of \cite{ansari2021mentalhealth}, who also used EDA and Back Translation as baseline approaches for their medical data augmentation approach.

\begin{table}[H]
\fontsize{9}{10}\selectfont
\captionsetup{skip=6pt}
\centering
\caption{Comparison for the choice of different data augmentation approaches across multiple baseline models using the CHIP-CDN dataset.}
\begin{tabular}{lccccc}
\toprule
DA Approaches & BiLSTM & BERT-base & CDN-Baseline & B\textsc{i-hard}NCE & B\textsc{i-hard}NCE \\
(Metric) & (Acc) & (Acc) & (F1) & (RECALL@5) & (NDCG@5) \\
\midrule
None & 0.455 & 0.558 & 0.554 & 0.857 & 0.816\\
EDA & 0.451 & 0.519 & 0.561 & 0.795 & 0.798\\
BT & 0.466 & 0.556 & 0.578 & 0.845 & 0.828\\
DDA (ours) & \textbf{0.518} & \textbf{0.579} & \textbf{0.592} & \textbf{0.866} & \textbf{0.840}\\
\bottomrule
\end{tabular}
\label{CDN_result}
\end{table}

As shown in Table \ref{CDN_result}, both EDA and back-translation have a detrimental impact on performance in certain scenarios (especially EDA), but DDA enhances performance across all scenarios. 
An intuitive explanation of this phenomenon is that general data augmentation methods have the potential to alter the meaning of disease names significantly. For example, if random deletion \cite{wei2019eda} is applied to ``\begin{CJK*}{UTF8}{gbsn}阻塞性睡眠呼吸暂停\end{CJK*} (Obstructive Sleep Apnoea)", it can result in ``\begin{CJK*}{UTF8}{gbsn}阻塞性睡眠\end{CJK*} (Obstructive Sleep)", which represents a completely different disease. As a result, the matching relationship between the unnormalized and standard disease names is lost. In contrast, our data augmentation method maintains this matching relationship, leading to enhanced performance.

% However, we design our data augmentation methods to tailor to disease names, which are classified as sampling-based methods in \cite{DA_survey}, making it suitable for creating stronger applications.

We notice that the performance improvement is more pronounced in the BiLSTM model compared to the BERT-based models. This could be attributed to the fact that the pre-trained language models already contain some similar knowledge, but our proposed approach can further enhance their performance, demonstrating the effectiveness of DDA. Furthermore, the consistent performance improvement across all scenarios indicates that DDA is well-suited for the task and can serve as a plug-and-play module, offering benefits to various baseline models with different training objectives.

\subsection{Ablation Study (RQ2)}
% 加一些对结果的解释，interpretation
We conduct further assessments to illustrate the effectiveness of each category of data augmentation method on all the baseline models. We assess their impact by removing each type of method one by one and observing the resulting performance. 
Similarly, we assess the impact of the semantic filtering module by removing the filtering rules one by one.
As shown in Table \ref{CDN_ablation}, the removal of data generated by either type of method led to a decline in performance. We also observe a performance decline when removing either of the filtering methods. This shows that all the data augmentation and filtering methods are effective.

\begin{table}[H]
\fontsize{9}{10}\selectfont
\captionsetup{skip=6pt}
\centering
\caption{Ablation study for the DDA approach. We remove our proposed data augmentation and semantic filtering methods one by one and evaluate the results.}
\begin{tabular}{lccccc}
\toprule
Settings & BiLSTM & BERT-base & CDN-Baseline & B\textsc{i-hard}NCE & B\textsc{i-hard}NCE \\
(Metric) & (Acc) & (Acc) & (F1) & (RECALL@5) & (NDCG@5) \\
\midrule
DDA (full) & \textbf{0.518} & \textbf{0.579} & \textbf{0.592} & \textbf{0.866} & \textbf{0.840}\\
- AR & 0.487 & 0.568 & 0.588 & 0.861 & 0.833\\
- MGA & 0.455 & 0.558 & 0.554 & 0.857 & 0.816\\
\midrule
- ngm & 0.505 & 0.572 & 0.581 & 0.858 & 0.830\\
- similarity & 0.485 & 0.560 & 0.574 & 0.857 & 0.826\\
\bottomrule
\end{tabular}
\label{CDN_ablation}
\end{table}

\subsection{Smaller Datasets Experiments (RQ3)}

\begin{figure}[htbp]
    \centering
    \includegraphics[width=1.0\textwidth,height=0.32\textwidth]{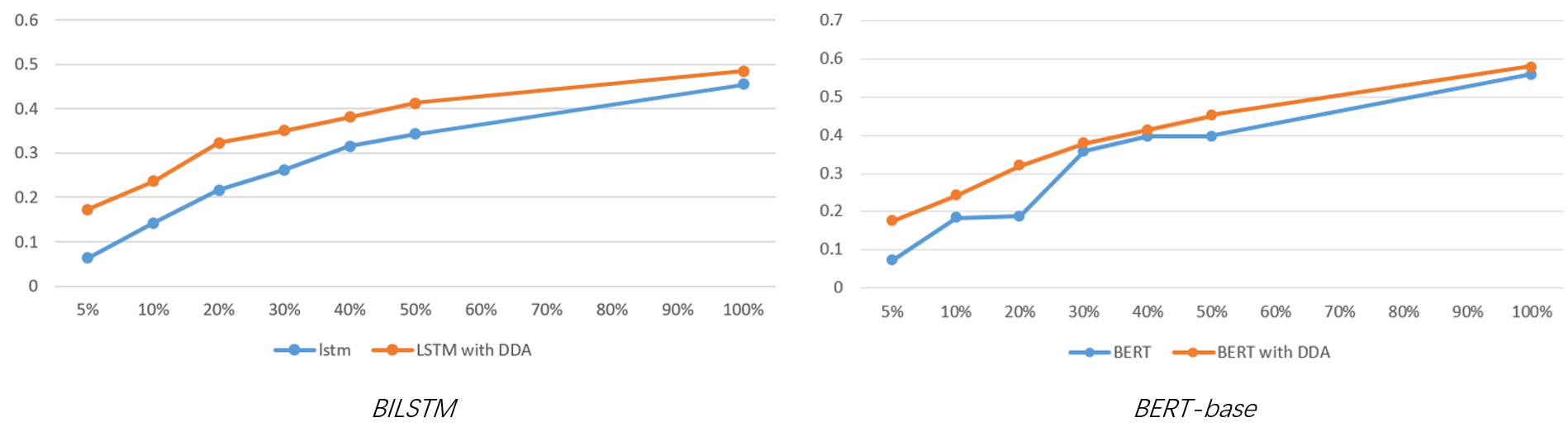}
    \caption{Performance comparison on smaller datasets for BiLSTM and BERT-base. The smaller datasets are derived by randomly sampling the original CHIP-CDN training set, and the validation set of CHIP-CDN stays the same.}
    \label{small_dataset}
\end{figure}

We are particularly interested in evaluating the performance improvements over smaller datasets derived from CHIP-CDN since the data scarcity problem is more severe in smaller datasets, and it can further validate our assumption that our approach gives the model comprehensive knowledge about the disease names and the classification system. Therefore, we conduct experiments to evaluate the scenario in which the training set size is restricted (from 5\% to 100\% of the original training set size). For the convenience of training, we only leverage standard disease names in the ICD system during data augmentation. No data from the disease name normalization training set is used.

\begin{table}[H]
\fontsize{9}{10}\selectfont
\captionsetup{skip=6pt}
\centering
\caption{Comparison between the performance of zero-shot inference and full fine-tuning over various baseline models.}
\begin{tabular}{lccccc}
\toprule
Settings & BiLSTM & BERT-base & CDN-Baseline & B\textsc{i-hard}NCE & B\textsc{i-hard}NCE \\
(Metric) & (Acc) & (Acc) & (F1) & (RECALL@5) & (NDCG@5) \\
\midrule
DDA (full) & 0.518 & 0.579 & 0.592 & 0.866 & 0.840\\
Zero-Shot & 0.034 & 0.073 & 0.113 & 0.672 & 0.670\\
\bottomrule
\end{tabular}
\label{zero-shot_performance}
\end{table}

% 加一些interpretation
The performance gap between whether to use our data augmentation or not is significantly larger when fewer training data is used, as depicted in Figure \ref{small_dataset}. When the size of the training set increases, both curves steadily increase. We also notice that the performance gain is higher when the size of the training set is smaller. We further perform zero-shot inferences for all four baseline models, where the inference is conducted without fine-tuning the models. The comparison results between zero-shot and full fine-tuning are shown in Table \ref{zero-shot_performance}. It is particularly evident that B\textsc{i-hard}NCE is able to recover nearly 80\% of the full performance for RECALL@5 and NDCG@5 in zero-shot settings. All the results above shows that the model has learned the general knowledge about the disease name system as expected.

\subsection{Comparisons with LLM Baselines (RQ4)}
In the field of Natural Language Processing, the Large Language Models (LLMs) have been developing very fast. LLMs can solve various problems by generating natural language and have demonstrated the ability to perform incredibly well on a wide range of tasks. This raises a crucial question: Can the existing approach for the disease name normalization task be replaced by LLMs? To answer this question, we compare the task performance between LLM-based approaches and our proposed approach.

We choose to use CDN-Baseline with DDA to compare with three LLM baselines: ChatGPT, GPT-4, and ChatGLM. We use the results reported in \cite{zhu2023promptcblue} for the LLM baselines. They also address the task using the ``recall-match" two-step fashion, where BM25 \cite{bm25} is the method to recall all the relevant disease names, and then the LLMs are prompted to select (match) the final answer from the retrieved names. The reason why they use such a pipeline is likely because it is hard to make LLMs directly perform the classification problem with a large label space (in this case, 40,474).
We compare them with CDN-Baseline with DDA because they not only share the pipeline but also the F1 evaluation metric.

\begin{figure}[htbp]
    \centering
    \includegraphics[width=0.65\textwidth]{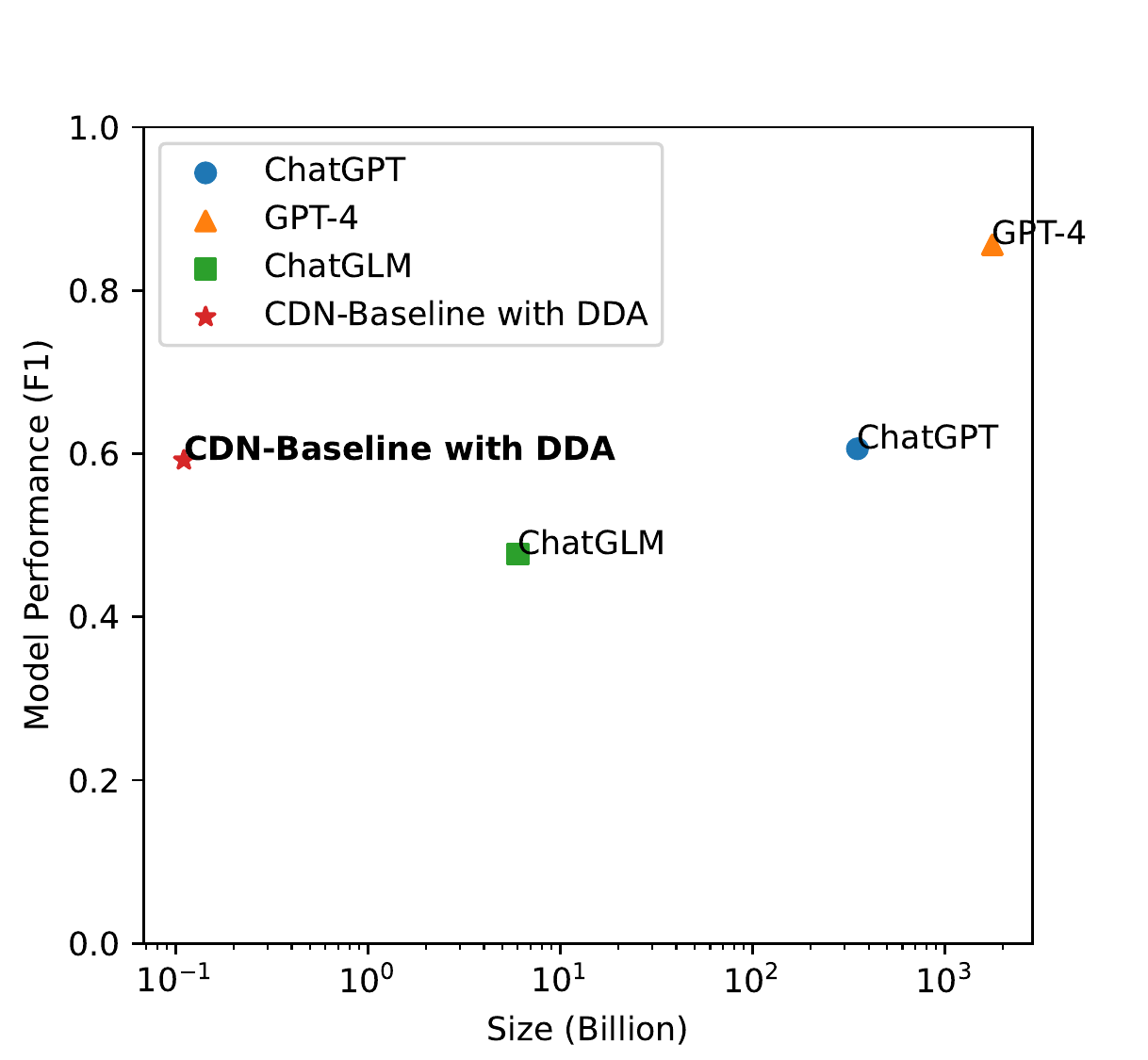}
    \caption{Performance comparison between LLM baselines and our proposed approach (CDN-Baseline with DDA).}
    \label{llm_comparison}
\end{figure}

% 最后分析结果，给出解释：为什么LLM只做了ranking
We visualize the results by comparing the model size of the task performance in Figure \ref{llm_comparison}. The size of ChatGPT and GPT-4 are estimated based on online sources (\cite{GPT3.5size, GPT4size}). 
The result demonstrates our approach has the best tradeoff between model performance and model size. Specifically, our proposed approach achieves on par performance with ChatGPT, despite our model size being over 3,000 times smaller. Our approach can significantly outperform a model over 50 times larger in size (ChatGLM 6B vs. CDN-Baseline with DDA 110M).

\section{Conclusion}
In this work, we investigate the critical task of disease name normalization in Chinese, a process essential for intelligent healthcare applications such as consultation, auxiliary diagnosis, and ICD coding. We identify the primary challenge of this task to be the scarcity of labeled data for model training. To address the issue, we introduce a novel data augmentation approach comprising two categories of data augmentation methods and some supporting modules. Our data augmentation methods involve Axis-word Replacement (AR) and Multi-Granularity Aggregation (MGA), which generate new training pairs by manipulating key elements of disease names and aggregating based on the hierarchical nature of disease classifications in the ICD system. We demonstrate through experiments that, unlike general text augmentation approaches, our approach significantly enhances performance across various baseline models for the Chinese disease name normalization task. It also achieves the best tradeoff between performance and model size when compared with LLM baselines. Our findings suggest that our data augmentation approach can serve as a robust tool to mitigate data scarcity of the disease name normalization task.

\section{Limitation and Future Work}
While the proposed approach has shown effectiveness, it still has some limitations. Firstly, there is no guarantee that the generated disease names are authentic, which could introduce biases to the model due to misinformation. We attempt to address this issue by utilizing the pretrain-finetune paradigm, but this approach does not fully resolve the problem. Secondly, although we believe it is possible, it remains unclear whether the approach can be effectively applied to English disease names. We have observed that in English disease names, a single word may represent multiple types of axis words, making it challenging to adapt the concept in English. An example is that ``Pylephlebitis" incorporates not only the meaning of portal vein but also the meaning of inflammation. Furthermore, conducting such experiments requires a high-quality disease name normalization dataset in English.

In this work, we have demonstrated the effectiveness of our DDA approach. However, we have not conducted a theoretical analysis to elucidate the underlying mechanisms that contribute to its effectiveness. Therefore, our future research will focus on exploring these mechanisms. Additionally, to further prevent the injection of misinformation, we plan to develop loss function terms in our future work that will enable the effective selection of more valuable data from the results of the data augmentation module.




\bibliographystyle{elsarticle-num} 
\bibliography{references}





% \end{thebibliography}

\appendix
\section{Data Augment Result Statics}
\label{sec:DA result statistic}
The number of data pairs generated by each data augmentation method in our proposed approach are as follows:
\begin{itemize}
    \item AR1: 332231
    \item AR2: 48857
    \item MGA - Code: 32145
    \item MGA - Region: 6239
\end{itemize}

% Table \ref{augument data result} shows the statistical results of the data generated by MGA and AR data augmentation methods.

% \begin{table}[H]
% % \fontsize{9}{10}\selectfont
% \captionsetup{skip=6pt}
% \centering
% \caption{The number of data pairs in each method generated by our data augmentation methods.}
% \begin{tabular}{lccccccccc}
% \toprule
% Methods& AR1 & AR2 & MGA - Code & MGA - Region \\
% \midrule
% Num &  332231 & 48857 & 32145 & 6239 \\
% \bottomrule
% \end{tabular}
% \label{augument data result}
% \end{table}

\section{Hyperparameter Settings}
Table \ref{Hyperparameter settings} shows the hyperparameter settings of our choices. For models that randomly initialize their parameters like BilSTM \cite{lstm}, it is possible to set a large learning rate and a large number of iterations to ensure adequate training. However, for models that rely on a pre-trained model checkpoint such as BERT \cite{devlin2018bert} as the backbone, we observe that setting a small learning rate and a small number of training iterations can lead to improved performance, likely due to its ability to mitigate catastrophic forgetting of knowledge within the original checkpoint.

\begin{table}[H]
% \fontsize{9}{10}\selectfont
\captionsetup{skip=6pt}
\centering
\caption{Hyperparameter settings for all the baseline models}
\begin{tabular}{lcccc}
\toprule
Model & Stage & Batch Size & Learning Rate & Epoch \\
\midrule
BiLSTM       & Pre-training & 256 & 1e-3          & 100    \\
BiLSTM       & Fine-tuning  & 64 & 1e-3          & 100    \\
BERT         & Pre-training & 256 & 1e-5           & 10      \\
BERT         & Fine-tuning  & 64 & 1e-4          & 100      \\
CDN-Baseline & Pre-training & 256 & 5e-6           & 1      \\
CDN-Baseline & Fine-tuning  & 64 & 5e-5          & 3     \\
B\textsc{i-hard}NCE & Pre-training & 16 & 3e-5      & 1      \\
B\textsc{i-hard}NCE & Fine-tuning & 16 & 3e-5      & 10     \\
\bottomrule
\end{tabular}
\label{Hyperparameter settings}
\end{table}

\section{Examples for each data augmentation technique}
Table \ref{DA_technique_examples} gives an example for every data augmentation technique.

\begin{table}[H]
% \fontsize{10}{12}\selectfont
\captionsetup{skip=6pt}
\centering
\caption{Examples of the data generated by each data augmentation technique. The ``find" and ``replace" operations correspond to the operation illustrated in Figure \ref{methods}.}
\begin{tabular}{p{4cm}p{8.7cm}}
\toprule
Technique & Example \\
\midrule % \begin{CJK*}{UTF8}{gbsn}髂总动脉夹层\end{CJK*}
\multirow{2}{4cm}{AR1 - Region} & Find: \begin{CJK*}{UTF8}{gbsn}踝关节骨折脱位\end{CJK*} $\rightarrow$ \begin{CJK*}{UTF8}{gbsn}腰椎骨折\end{CJK*} \\
& Replace: \begin{CJK*}{UTF8}{gbsn}腰椎骨折脱位\end{CJK*} $\rightarrow$ \begin{CJK*}{UTF8}{gbsn}腰椎骨折\end{CJK*} \\
\midrule
\multirow{2}{4cm}{AR1 - Center} & Find: \begin{CJK*}{UTF8}{gbsn}左内踝关节囊肿\end{CJK*} $\rightarrow$ \begin{CJK*}{UTF8}{gbsn}踝关节骨折\end{CJK*} \\
& Replace: \begin{CJK*}{UTF8}{gbsn}左内踝关节骨折\end{CJK*} $\rightarrow$ \begin{CJK*}{UTF8}{gbsn}踝关节骨折\end{CJK*} \\
\midrule
\multirow{2}{4cm}{AR1 - Characteristic} & Find: \begin{CJK*}{UTF8}{gbsn}重度慢性牙周炎\end{CJK*} $\rightarrow$ \begin{CJK*}{UTF8}{gbsn}急性牙周炎\end{CJK*} \\
& Replace: \begin{CJK*}{UTF8}{gbsn}重度急性牙周炎\end{CJK*} $\rightarrow$ \begin{CJK*}{UTF8}{gbsn}急性牙周炎\end{CJK*} \\
\midrule
\multirow{2}{4cm}{AR2 - Region} & Find: \begin{CJK*}{UTF8}{gbsn}踝关节骨折脱位\end{CJK*} $\rightarrow$ \begin{CJK*}{UTF8}{gbsn}踝关节骨折\end{CJK*} \\
& Replace: \begin{CJK*}{UTF8}{gbsn}腰椎骨折脱位\end{CJK*} $\rightarrow$ \begin{CJK*}{UTF8}{gbsn}腰椎骨折\end{CJK*} \\
\midrule
\multirow{2}{4cm}{AR2 - Center} & Find: \begin{CJK*}{UTF8}{gbsn}左内踝关节囊肿\end{CJK*} $\rightarrow$ \begin{CJK*}{UTF8}{gbsn}踝关节囊肿\end{CJK*} \\
& Replace: \begin{CJK*}{UTF8}{gbsn}左内踝关节骨折\end{CJK*} $\rightarrow$ \begin{CJK*}{UTF8}{gbsn}踝关节骨折\end{CJK*} \\
\midrule
\multirow{2}{4cm}{AR2 - Characteristic} & Find: \begin{CJK*}{UTF8}{gbsn}重度慢性牙周炎\end{CJK*} $\rightarrow$ \begin{CJK*}{UTF8}{gbsn}慢性牙周炎\end{CJK*} \\
& Replace: \begin{CJK*}{UTF8}{gbsn}重度急性牙周炎\end{CJK*} $\rightarrow$ \begin{CJK*}{UTF8}{gbsn}急性牙周炎\end{CJK*} \\
\midrule
\multirow{1}{4cm}{MGA - Code 1} & \begin{CJK*}{UTF8}{gbsn}急性脑膜炎症\end{CJK*} $\rightarrow$ \begin{CJK*}{UTF8}{gbsn}脑膜炎\end{CJK*} \\
\midrule
\multirow{1}{4cm}{MGA - Code 2} & \begin{CJK*}{UTF8}{gbsn}急性脑膜炎\end{CJK*} $\rightarrow$ \begin{CJK*}{UTF8}{gbsn}脑膜炎\end{CJK*} \\
\midrule
\multirow{1}{4cm}{MGA - Region 1} & \begin{CJK*}{UTF8}{gbsn}副乳腺恶性肿瘤\end{CJK*} $\rightarrow$ \begin{CJK*}{UTF8}{gbsn}乳腺恶性肿瘤\end{CJK*} \\
\midrule
\multirow{1}{4cm}{MGA - Region 2} & \begin{CJK*}{UTF8}{gbsn}右乳房乳腺恶性肿瘤\end{CJK*} $\rightarrow$ \begin{CJK*}{UTF8}{gbsn}乳腺恶性肿瘤\end{CJK*} \\
\bottomrule
\end{tabular}
\label{DA_technique_examples}
\end{table}
% 在caption里解释一下find和change，说跟figure x里的对应

% \begin{table}[H]
% % \fontsize{9}{10}\selectfont
% \captionsetup{skip=6pt}
% \centering
% \begin{tabular}{lc}
% \toprule
% Technique & Example \\
% \midrule
% AR1 - Location       & Pre-training \\
% AR1 - Center       & Pre-training \\
% AR1 - Characteristics         & Pre-training \\
% AR2 - Location         & Pre-training \\
% AR2 - Center          & Pre-training \\
% AR2 - Characteristics        & Pre-training \\
% MGA - Code 1         & Pre-training \\
% MGA - Code 2        & Pre-training \\
% MGA - Location 1         & Pre-training \\
% MGA - Location 2        & Pre-training \\
% \bottomrule
% \end{tabular}
% \caption{All the proposed data augmentation techniques, accompanied by an example for each technique.}
% \label{DA_technique_examples}
% \end{table}

\section{Pseudo-code}
\label{apdx:pseudo-code}
In this section, we present the pseudo-code for the four proposed data augmentation methods.

\begin{table}[H]
% \fontsize{9}{10}\selectfont
\captionsetup{skip=6pt}
\centering
\caption{Annotations used in the algorithms. Note that ``disease names" can represent both unnormalized disease names or standard disease names.}
% \footnotesize
\resizebox{\textwidth}{!}{
\begin{tabular}{lc}
\toprule
Descriptions & Notations \\
\midrule
Axis word & $a1$, $a2$, $a3$, etc. \\
List of axis words & $A1$, $A2$, $A3$, etc. \\
Axis type - Disease Center & $dce$ \\
Axis type - Anatomical Region & $al$ \\
Axis type - Disease Characteristic & $dch$ \\
Larger region & $lar$ \\
List of shared axis words between two diseases & $SA$ \\
List of differing axis words between two diseases & $DiA$ \\
List of differing axis words in the first disease when comparing two diseases & $DiA1$ \\
List of differing axis words in the second disease when comparing two diseases & $DiA2$ \\
Unnormalized disease names (UDN) & $u1$, $u2$, $u3$, etc. \\
Standard disease names (SDN) & $s1$, $s2$, $s3$, etc. \\
Disease names (can be either a UDN or an SDN) & $d1$, $d2$, $d3$, etc. \\
\bottomrule
\end{tabular}
}
\label{tab:concepts_in_algorithms}
\end{table}

% AR1
\begin{algorithm}
\caption{Axis-word Replacement 1 (AR1)}
\label{algorithm:AR1}
\begin{algorithmic}[1]
\State \textbf{Input:} 
\Statex \hspace{\algorithmicindent}$training\_set$ - List of disease pairs from the disease name 
\Statex \hspace{\algorithmicindent} normalization training set.
\Statex \hspace{\algorithmicindent}$ICD\_list$ - The standard ICD system.
\State \textbf{Output:} $augmented\_pairs$ - List of augmented disease pairs.
\Procedure{AR1}{$training\_set, ICD\_list$}
    \State $augmented\_pairs \gets []$
    \For{each $d1$ in $(training\_set \cup ICD\_list)$}
        \State $A1 \gets NER(d1)$
        \For{each $s1$ in $ICD\_list$}
            \State $A2 \gets NER(s1)$
            \State $SA, DiA1, DiA2 \gets comparing\_axis\_words(A1, A2)$
            \If{$\text{len}(SA) \neq 0$ and len$(DiA1)$ = len$(DiA2) = 1$}
                \State $d2 \gets d1.replace\_axis(DiA1[0], DiA2[0])$
                \State $augmented\_pairs.append((d2, s1))$
            \EndIf
        \EndFor
    \EndFor
    \State \Return $augmented\_pairs$
\EndProcedure
\end{algorithmic}
\end{algorithm}

% AR2
\begin{algorithm}
\caption{Axis-word Replacement 2 (AR2)}
\label{algorithm:AR2}
\begin{algorithmic}[1]
\State \textbf{Input:} 
\Statex \hspace{\algorithmicindent}$training\_set$ - List of disease pairs from the disease name 
\Statex \hspace{\algorithmicindent} normalization training set.
\Statex \hspace{\algorithmicindent}$ICD\_list$ - The standard ICD system.
\State \textbf{Output:} $augmented\_pairs$ - List of augmented disease pairs.
\Procedure{AR2}{$training\_set, ICD\_list$}
    \State $augmented\_pairs \gets []$
    \For{each $(u1, s1)$ in $training\_set$}
        \State $A1 \gets NER(u1)$
        \State $A2 \gets NER(s1)$
        \If{$A1 = A2$}
            \For{each $s2$ in $ICD\_list$}
                \State $A3 \gets NER(s2)$
                \State $SA, DiA1, DiA2 \gets comparing\_axis\_words(A2, A3)$
                \If{len$(A2)=$len$(A3)$ and len$(DiA1)$=len$(DiA2)=1$}
                    \State $s3 \gets s1.replace\_axis(DiA1[0], DiA2[0])$
                    \State $u2 \gets u1.replace\_axis(DiA1[0], DiA2[0])$
                    \State $augmented\_pairs.append((u2, s3))$
                \EndIf
            \EndFor
        \EndIf
    \EndFor
    \State \Return $augmented\_pairs$
\EndProcedure
\end{algorithmic}
\end{algorithm}

% MGA-code
\begin{algorithm}
\caption{Multi-Granularity Aggregation - Code (MGA-Code)}
\label{algorithm:MGA-Code}
\begin{algorithmic}[1]
\State \textbf{Input:} 
\Statex \hspace{\algorithmicindent}$training\_set$ - List of disease pairs from the disease name 
\Statex \hspace{\algorithmicindent} normalization training set.
\Statex \hspace{\algorithmicindent}$ICD\_list$ - The standard ICD system.
\State \textbf{Output:} $augmented\_pairs$ - List of augmented disease pairs.
\Procedure{MGA-Code}{$training\_set, ICD\_list$}
    \State $augmented\_pairs \gets []$
    \For{each $d1$ in $(training\_set \cup ICD\_list)$}
        \If{$len($ICD-code$(d1)) = 6$}
            \State $code_6 = $ICD-code$(d1)$
            \State $code_4 = code_6[0:3]$ // Extract the first four digits
            \State $s1 = $map\_disease$(code_4)$
            \State $augmented\_pairs.append((d1, s1))$
        \EndIf
    \EndFor
    \State \Return $augmented\_pairs$
\EndProcedure
\end{algorithmic}
\end{algorithm}

% MGA-position
\begin{algorithm}
\caption{Multi-Granularity Aggregation - Region (MGA-Region)}
\label{algorithm:MGA-Position}
\begin{algorithmic}[1]
\State \textbf{Input:} 
\Statex \hspace{\algorithmicindent}$training\_set$ - List of disease pairs from the disease name 
\Statex \hspace{\algorithmicindent} normalization training set.
\Statex \hspace{\algorithmicindent}$ICD\_list$ - The standard ICD system.
\State \textbf{Output:} $augmented\_pairs$ - List of augmented disease pairs.
\Procedure{MGA-Region}{$training\_set, ICD\_list$}
    \State $augmented\_pairs \gets []$
    \For{each $d1$ in $(training\_set \cup ICD\_list)$}
        \State $A1 \gets NER(d1)$
        \For{each $s1$ in $ICD\_list$}
            \State $A2 \gets NER(s1)$
            \State $SA, DiA1, DiA2 \gets comparing\_axis\_words(A1, A2)$
            \If{len$(SA) \geq 1$ and len$(DiA1)$=len$(DiA2)=1$}
                \If{$type(DiA1)=al$ and $DiA2[0]$=$DiA1[0].lar$}
                    \State $augmented\_pairs.append((d1, s1))$
                \EndIf
            \EndIf
        \EndFor
    \EndFor
    \State \Return $augmented\_pairs$
\EndProcedure
\end{algorithmic}
\end{algorithm}

\end{document}